\documentclass{article}

\usepackage[preprint,nonatbib]{nips_2018}

\usepackage[utf8]{inputenc} 
\usepackage[T1]{fontenc}    
\usepackage{hyperref}       
\usepackage{url}            
\usepackage{booktabs}       
\usepackage{amsfonts}       
\usepackage{microtype}      
\usepackage{subfigure}
\usepackage{epsfig}
\usepackage{graphicx}
\usepackage{amsmath}
\usepackage{amssymb}
\usepackage{subfigure}
\usepackage{float}

\title{MicronNet: A Highly Compact Deep Convolutional Neural Network Architecture for Real-time Embedded Traffic Sign Classification}

\author{
  Alexander Wong$^{1}$, Mohammad Javad Shafiee$^{1}$, and Michael St. Jules$^{1,2}$\\
  $^{1}$Waterloo Artificial Intelligence Institute, University of Waterloo, Waterloo, ON, Canada\\
  $^{2}$DarwinAI Corp., Waterloo, ON, Canada\\
  \texttt{a28wong@uwaterloo.ca} \\
}

\begin{document}

\maketitle

\begin{abstract}
Traffic sign recognition is a very important computer vision task for a number of real-world applications such as intelligent transportation surveillance and analysis.  While deep neural networks have been demonstrated in recent years to provide state-of-the-art performance traffic sign recognition, a key challenge for enabling the widespread deployment of deep neural networks for embedded traffic sign recognition is the high computational and memory requirements of such networks.  As a consequence, there are significant benefits in investigating compact deep neural network architectures for traffic sign recognition that are better suited for embedded devices.  In this paper, we introduce MicronNet, a highly compact deep convolutional neural network for real-time embedded traffic sign recognition designed based on macroarchitecture design principles (e.g., spectral macroarchitecture augmentation, parameter precision optimization, etc.)  as well as numerical microarchitecture optimization strategies.  The resulting overall architecture of MicronNet is thus designed with as few parameters and computations as possible while maintaining recognition performance, leading to optimized information density of the proposed network.  The resulting MicronNet possesses a model size of just $\sim$1MB and $\sim$510,000 parameters ($\sim$27x fewer parameters than state-of-the-art) while still achieving a human performance level top-1 accuracy of 98.9\% on the German traffic sign recognition benchmark.  Furthermore, MicronNet requires just $\sim$10 million multiply-accumulate operations to perform inference, and has a time-to-compute of just 32.19 ms on a Cortex-A53 high efficiency processor.  These experimental results show that highly compact, optimized deep neural network architectures can be designed for real-time traffic sign recognition that are well-suited for embedded scenarios.
\end{abstract}
		\section{Introduction}
		
				\begin{figure}
					\begin{center}
							\includegraphics[width = 8.5 cm]{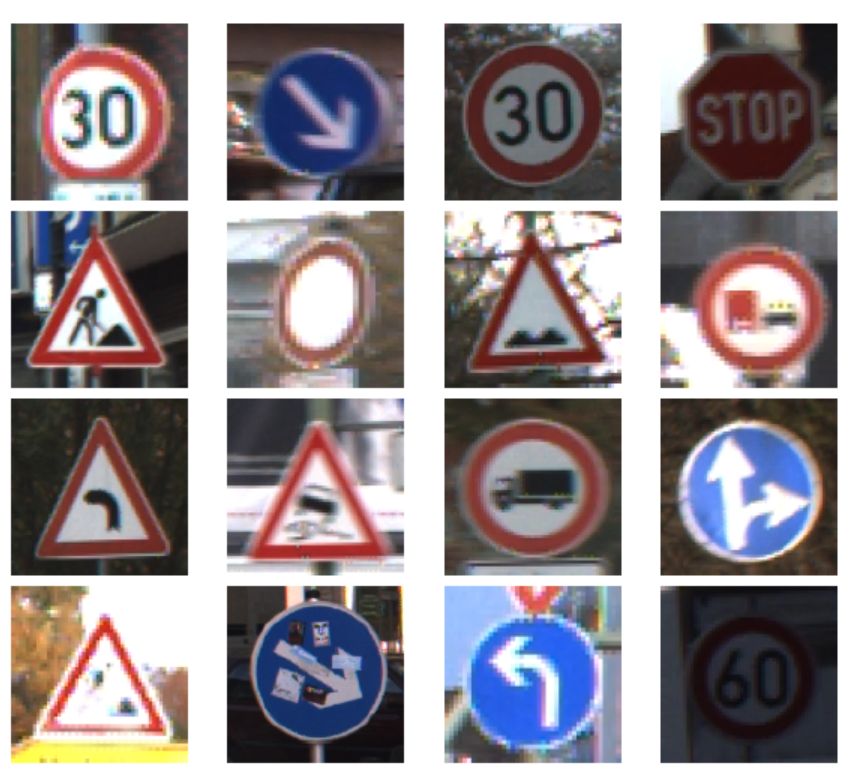}
						\caption{The goal of the traffic signal recognition problem is to identify which type of traffic sign is in the scene.  For context, some example images of traffic signs from the German traffic sign recognition benchmark~\cite{GTSRB} are shown.   }
						\label{fig:signsamples}
					\end{center}
				\end{figure}
		
		Traffic sign recognition can be considered an important computer vision task for a number of real-world applications such as intelligent transportation surveillance and analysis (see Figure~\ref{fig:signsamples}).  The arrival of modern breakthroughs in deep learning~\cite{lecun2015deep,krizhevsky2012imagenet} has resulted in significant state-of-the-art results for traffic sign recognition, with much of the research focused on designing deep convolutional neural networks for improved accuracy~\cite{Sermanet,Ciresan2011,Ciresan2012,Wu2013,Jin2014,Aghdam,ArcosGarcia2018,Aghdam2017}.

Despite the fact that such traffic sign recognition networks have shown state-of-the-art object detection accuracies beyond what can be achieved by previous state-of-the-art methods, a key challenge for enabling the widespread deployment of deep neural networks for embedded traffic sign recognition is the high computational and memory requirements of such networks.  For example, the committee of deep convolutional neural networks proposed by Ciresan et al.~\cite{Ciresan2012} consists of $\sim$38.5 million parameters while the ensemble of deep convolutional neural networks trained via hinge loss as proposed by Jin et al.~\cite{Jin2014} consists of $\sim$23.2 million parameters.  More recently, the state-of-the-art deep convolutional network with spatial transformers proposed by Arcos-Garcia et al.~\cite{ArcosGarcia2018}, while having fewer parameters than the aforementioned approaches, still consisted of over $\sim$14 million parameters.  At a significantly smaller sizes than the aforementioned configurations, the macroarchitectures proposed by Ciresan et al.~\cite{Ciresan2012} still consist of $\sim$1.5 million parameters.  As such, the design of more compact and efficient deep neural network architectures for traffic sign recognition is highly desired for embedded applications.

		Recently, there has been an increasing focus in exploring small deep neural network architectures that are more suitable for embedded devices~\cite{howard2017mobilenets,iandola2016squeezenet,sandler2018,Hasanpour,Aghdam,Wong,shafiee2017squishednets,wu2016squeezedet,tinyyolo}.  For example, in the work by Iandola et al.~\cite{iandola2016squeezenet}, three key design strategies  were leveraged to create compact macroarchitectures: 1) filter quantity  reduction, 2) input channel reduction, and 3) late downsampling in the network.  As a result of such design strategies, a compact SqueezeNet macroarchitecture was introduced that comprised of Fire modules that was $\sim$50X smaller than AlexNet with comparable accuracy on ImageNet~\cite{imagenet_cvpr09} for 1000 classes.  In the work by Howard et al.~\cite{howard2017mobilenets}, they leveraged depth-wise separable convolutions to reduce the number of parameters, as well as two global hyperparameters based on network width and resolution for finding the tradeoff between latency and accuracy.  Sandler et al.~\cite{sandler2018} expanded upon this by introducing an inverted residual structure that enabled further reductions in the number of parameters while maintaining high performance.  Aghdam et al.~\cite{Aghdam} presented techniques for optimizing the efficiency of deep neural network architectures for the specific purpose of traffic sign recognition.  Based on the practical principles they discussed for building small deep neural network architectures for traffic sign recognition, the authors were able to create a high-performance deep neural network consisting of just $\sim$1.74 million parameters, while still achieving great accuracy.
		
		In this study, we introduce \textbf{MicronNet}, a highly compact deep convolutional neural network designed specifically for real-time embedded traffic sign recognition.  In MicronNet's highly optimized network architecture, the underlying microarchitecture of each convolutional layer in the network (with microarchitecture here referring to the number and size of convolutional filters) is numerically optimized to have as few parameters and computations as possible while maintaining recognition performance, hence resulting in an optimized information density for the underlying network.  Furthermore, the macroarchitecture of the proposed MicronNet network is designed via macroarchitecture design strategies (e.g., spectral macroarchitecture augmentation, parameter precision optimization, etc.) that encourage improved computational efficiency and efficacy in embedded environments.  As such, the main contribution of this work is the investigation and exploration of integrating design principles and optimization strategies at both the microarchitecture level and the macroarchitecture level to design deep neural networks with optimized information densities that satisfy real-time embedded requirements while achieving strong accuracy, thus enabling real-time embedded traffic sign recognition.
		
		This paper is organized as follows.  Section 2 describes the highly optimized network architecture and design considerations underlying the proposed MicronNet network.  Section 3 presents experimental results that evaluate the efficacy of the proposed MicronNet network for real-time embedded traffic sign recognition, along with a discussion on some key observations about the network.  Finally, conclusions are drawn in Section 4.

		\section{Network architecture of MicronNet}
		
			\begin{figure}[t]
					\begin{center}
			\includegraphics[width = 8.5 cm]{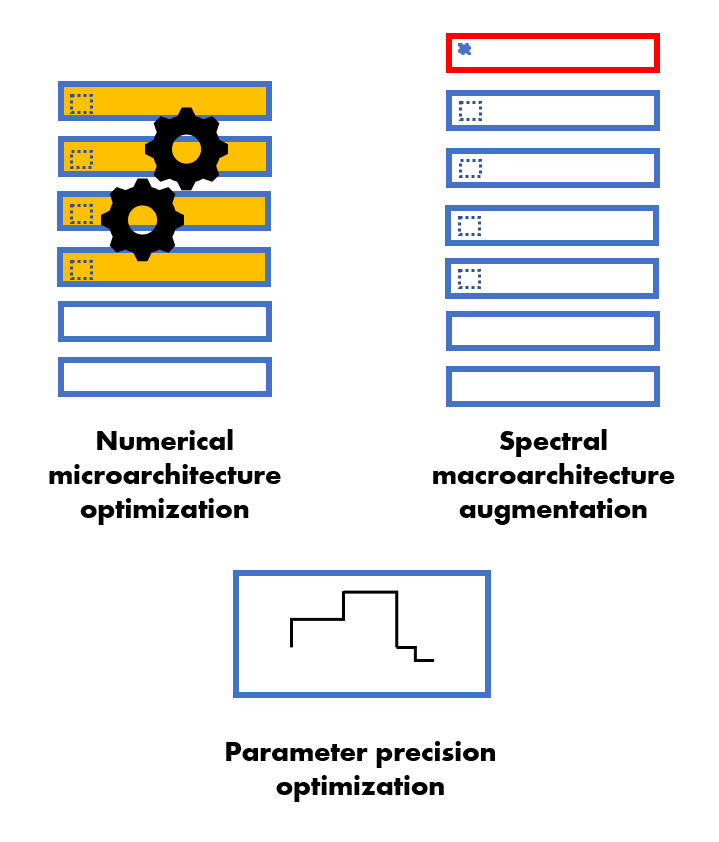}
			\caption{Integrated microarchitecture-level and macroarchitecture-level design principles and optimization strategies leveraged for designing MicronNet for high efficiency while maintaining strong accuracy.}
					\end{center}
			\label{fig:overview}
		\end{figure}	

		Leveraging macroarchitecture design principles such as those from~\cite{howard2017mobilenets,iandola2016squeezenet,sandler2018,Aghdam} and a numerical microarchitecture optimization strategy inspired by~\cite{Wong}, the overall network architecture of the proposed MicronNet network for real-time embedded traffic sign recognition is inspired by the network macroarchitecture described in~\cite{Aghdam} and takes the following microarchitecture-level and macroarchitecture-level design considerations and optimization strategies into account to greatly improve the efficiency of the resulting deep convolutional neural network while maintaining strong accuracy (see Figure~\ref{fig:overview}):
		
		\begin{itemize}
			\item Optimizing microarchitectures of each convolutional layer via numerical optimization for reduced number of parameters
			\item Incorporating spectral augmentations to produce a spectral-spatial macroarchitecture that further reduces number of parameters and computational complexity while maintaining strong accuracy
			\item Optimizing parameter precision for reduced model size while maintaining strong accuracy
		\end{itemize}
		
		Table~\ref{tab:micronNet} shows the overall architecture of the proposed MicronNet network architecture.   The proposed MicronNet network architecture is a 16-bit floating-point deep convolutional neural network composed of four convolutional layers, followed by two fully-connected layers and a softmax layer.  A combination of 1$\times$1 point-wise convolutional layer with 5$\times$5 and 3$\times$3 spatial convolutional layers form a spectral-spatial macroarchitecture for reducing complexity while maintaining accuracy.  Furthermore, rectified linear unit (ReLU) activation functions are leveraged within the proposed MicronNet network architecture for low computational complexity and better suitability for real-time embedded applications.  Each of the design considerations in the design of MicronNet is discussed in detail below.

		\subsection{Numerical microarchitecture optimization}

		\begin{table}
			\begin{center}
\vspace{1 cm}
				\caption{The optimized network architecture underlying MicronNet}
				\label{tab:micronNet}
				\normalsize
				\begin{tabular}{l|l|c}
					\hline \hline
					Type / Stride / Pad    & Filter Shape       & Input Size \\ \hline
					Conv / s1 / p0        & $ 1\times1 \times 1$  & $48 \times 48$\\ \hline
					Conv / s1 / p0        & $ 5\times5 \times 29$  & $48 \times 48$\\ \hline		
					Pool / s2 / p0       & $3 \times 3$ & maxpool\\ \hline	
					Conv / s1 / p0        & $ 3\times3 \times 59$  & $22 \times 22$\\ \hline
					Pool / s2 / p0       & $3 \times 3$ & maxpool\\ \hline		
					Conv  / s1 / p0       & $ 3\times3 \times 74$  & $10 \times 10$\\ \hline	
					Pool / s2 / p0       & $3 \times 3$ & maxpool\\ \hline			
					FC / s1              & $ 1\times300$           & $1 \times 1184$\\ \hline	
					FC / s1               & $ 1\times300$           & $1 \times 300$\\ \hline	
					Softmax / s1               & Classifier           & $1 \times 43$\\ \hline		
				\end{tabular}
			\end{center}
		\end{table}

		The first design consideration in obtaining an ideal network architecture for real-time embedded traffic sign recognition in this study is to optimize the network microarchitecture of the proposed MicronNet network.  One of the key challenges to identifying the ideal microarchitecture for each of the individual convolutional layers in the deep neural network is to achieve a fine balance between modeling performance and model size as well as computations involved.  While a number of existing techniques have focused on uniform microarchitecture design~\cite{howard2017mobilenets,sandler2018}, the strategy employed here is instead focused on numerical microarchitecture optimizations that operates at a more fine-grain level than other techniques, as it was found by the authors to yield a better balance between modeling performance and model size as well as reduced computations.
		
		Taking that mentality into account here, the key design parameters of the microarchitectures of each convolutional layer are the number of convolutional filters that form the microarchitecture, and their associated sizes.  Therefore, here we optimize the number of convolutional filters and their associated sizes in each convolutional layer via numerical optimization.  More specifically, the key objective leveraged here is to minimize the number of parameters that compose each convolutional layer in the network architecture while maintaining the overall accuracy of the network.

One quantifiable assessment of the relative amount of accuracy a given deep neural network captures with respect to a fundamental building block (in this case, a parameter) that ties well with this key objective is the information density~\cite{Canziani} of a deep neural network.  By taking into account both model size and network performance by means of a single metric, information density provides a good representation of the network's ability to utilize its full modeling capacity.  Therefore, a deep neural network with a good balance of being smaller with fewer parameters yet still maintaining strong performance would be characterized by a higher information density, and hence higher information density indicates better network efficiency and is thus our desired outcome.

In this study, the numerical microarchitecture optimization strategy is framed as a constrained optimization problem, where the set of optimization parameters $F$ is set as the number of convolutional filters and their associated sizes in each convolutional layer for a given network $\mathcal{N}$, and the goal is to numerically determine the optimal $F$ that minimizes the total number of network parameters (denoted here as $p(\mathcal{N};F)$) for a given $F$, with the validation accuracy $a_v{(\mathcal{N})}$ constrained to being greater than or equal to an accuracy lower-bound of $l$ (set to 98.5\% in this study based on the performance of~\cite{Ciresan2012}):
\begin{equation}
F = \min_{F}~p(\mathcal{N};F)~~\textrm{subject~to}~~a_v{(\mathcal{N})} \geq l.
\label{form}
\end{equation}
		
			\begin{figure}[t]
					\begin{center}
			\includegraphics[width = 8.5 cm]{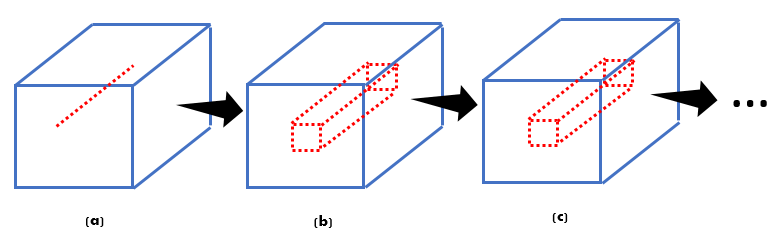}
\end{center}
			\caption{Spectral macroarchitecture augmentation: a 1$\times$1 point-wise convolutional layer (a) sits before a $5\times5$ convolutional layer (b) to form a spectral-spatial macroarchitecture where spectral features are first extracted through computing linear combinations of the input spectral channels in the point-wise convolutional layer, before spatial features are extracted in subsequent convolutional layers (c).}
			\label{fig:convolutions}
		\end{figure}	

An approximate solution to the above constrained optimization problem posed in Eq.~\ref{form} can be obtained using an iterative optimization approach.  The key advantage with leveraging such a numerical microarchitecture optimization strategy is that each layer has its own unique information density limits and thus the degree of microarchitecture optimization that can be achieved for each layer can differ substantially.  Therefore, a numerical microarchitecture optimization strategy allows significantly greater flexibility in obtaining the ideal microarchitectures for each convolutional layer with the optimal information densities without being constrained by the need for uniform fine-tuning.  As a result, the proposed MicronNet network architecture possesses highly optimized microarchitectures that is optimized for real-time embedded scenarios.

		\subsection{Spectral macroarchitecture augmentation}
		
		The second design consideration in obtaining an ideal network architecture for real-time embedded traffic sign recognition is to incorporate additional layers to the macroarchitecture of the MicronNet network that enable further reductions in computational complexity to be made while maintaining strong accuracy.  Taking inspiration from~\cite{howard2017mobilenets,iandola2016squeezenet,sandler2018} where 1$\times$1 convolutional layers are leveraged to reduce the number of parameters in the network while preserving modeling performance, we augment the proposed MicronNet network an additional 1$\times$1 convolutional layer placed at a strategic location where it would have the most impact on reducing computational complexity while having a positive impact on modeling performance.  More specifically, we take inspiration from work on spectral-spatial macroarchitectures such as that proposed in~\cite{Zhong}, which are designed to learn spectral features prior to learning spatial features in an end-to-end macroarchitecture, and incorporated an additional 1$\times$1 convolutional layer at the beginning of proposed MicronNet network architecture.  This 1$\times$1 convolutional layer sits before a $5\times5$ convolutional layer and acts as a pointwise feature transform layer where new features are built through computing linear combinations of the input spectral channels.  Therefore, from a theoretical perspective, one can view this 1$\times$1 convolutional layer as a spectral feature learning layer that learns the optimal spectral mixing projection between the input color channels in an image to produce a single-channel spectral feature map that feeds into subsequent convolutional layers (see Figure~\ref{fig:convolutions}), resulting in a spectral-spatial network macroarchitecture.  The key advantage of this additional 1$\times$1 convolutional layer compared to the strategy used by deep neural networks such as that proposed by~\cite{Aghdam}, which converts color input images into grayscale images using a pre-defined conversion scheme, is that it provides a much greater level of flexibility in learning a more discriminative spectral projection into a single feature channel than that can be achieved with a fixed grayscale conversion scheme.

  Based on empirical experiments, the augmentation of this additional pointwise convolutional layer to form a spectral-spatial network architecture enables us to greatly reduce the number of filters needed in the 5$\times$5 convolutional layer to obtain strong modeling accuracy.  In addition to reducing the number of parameters in the proposed MicronNet network, the reduction in the number of convolutional filters in the 5$\times$5 convolutional layer is very important as the convolutional filters are used to convolve inputs at the original image resolution, and as such reducing the number of convolutional filters result in a significant reduction in the number of computations that need to be performed.  It is important to note that this augmentation is performed on the proposed network architecture prior to the numerical microarchitecture optimization process.

		\subsection{Parameter Precision Optimization and Activation Function Selection}
		The third design consideration in obtaining an ideal network architecture for real-time embedded traffic sign recognition is to optimize the precision of the parameters used in the proposed MicronNet network.  For embedded applications, the computational requirements and memory requirements are typically quite strict and as such an effective strategy to address these requirements is to reduce the data precision of parameters in a deep neural network.  In particular, embedded processors often support accelerated mixed precision operations, and as a result leveraging such parameter precision considerations into the design of the deep neural network can result in noticeable improvements in computational time as well as memory storage for embedded scenarios.  For the MicronNet network architecture, all parameters are characterized with half precision floating-point data representations after training to enable further model size reductions while still achieving strong performance.  Alongside the use of fixed-point parameter precision for embedded applications, the utilization of half-precision floating-point parameter precision for deep neural networks has seen widespread adoption for embedded applications and hardware-accelerated in a wide range of embedded processors, including the Nvidia Tegra family of embedded processors as well as widely-used ARM embedded processors such as the Cortex-A53 high efficiency processor tested in this study.  In additional, we also produced a variant of the proposed MicronNet network architecture with 16-bit fixed-point data representation for comparison purposes.
		
Finally, to reduce the computational complexity of the proposed MicronNet network architecture, the rectified linear unit (ReLU) function is used as the activation function in the deep neural network since it is more suitable for real-time embedded applications when compared to other activation functions such as the scaled hyperbolic tangent function~\cite{LeCun} and the parametric rectifier linear unit (PReLU) function~\cite{He}.
		
		\subsection{Training}
		Here, we will discuss the training policy for learning the proposed MicronNet network.  The proposed MicronNet network was trained for 60,000 iterations in the Caffe framework with a training batch size of 50.  Stochastic gradient descent with momentum and exponential decay was utilized as the training policy with the base learning rate set to 0.007, the momentum set to 0.9, the learning rate decay step size set to 1000, and the learning rate decay rate set to 0.9996. A $l_2$ weight decay with rate 0.00001 was also used on the filters and matrices.

		\section{Experimental Results and Discussion}
		
		To study the efficacy of the proposed MicronNet for real-time embedded traffic sign recognition, we evaluate the following:
		\begin{itemize}
			\item Top-1 accuracy on the German traffic sign recognition benchmark (GTSRB)~\cite{GTSRB}
			\item Resource usage (model size, number of parameters, number of multiply-accumulate (MAC) operations, time-to-compute on a 1.2GHz Cortex-A53 high efficiency processor)
			\item Information density~\cite{Canziani} and NetScore~\cite{NetScore}
            \item Robustness against image degradation
		\end{itemize}
		
		For evaluation purposes, the following state-of-the-art traffic sign recognition networks were also compared:
		\begin{itemize}
			\item \textbf{STDNN~\cite{ArcosGarcia2018}}, deep convolutional neural network with spatial transformers,
			\item \textbf{HLSGD~\cite{Jin2014}}: hinge loss trained deep convolutional neural network,
			\item \textbf{MCDNN~\cite{Ciresan2012}}: multi-column deep convolutional neural network,
			\item \textbf{CDNN~\cite{Ciresan2012}}: Ciresan deep convolutional neural network.
		\end{itemize}
		
		\subsection{Dataset}
		The German traffic sign recognition benchmark (GTSRB)~\cite{GTSRB} used for evaluation purposes in this paper consists of color images of traffic signs (one traffic sign per image, with a total of 43 types of traffic signs) with image sizes varying from 15$\times$15 to 250$\times$250 pixels.  There are a total of 39,209 color images in the training set and a total of 12,630 images in the test set. To balance the number of samples in different classes as well as improve the generality of the resulting network, a number of different data augmentation techniques were leveraged including: i) rotation, ii) shifting, iii) sharpening, iv) Gaussian blur, v) motion blur, vi) HSV augmentation, and vii) mirroring.  As standard for evaluating performance using GTSRB, all images are cropped and all images are resized to 48$\times$48 pixels~\cite{Ciresan2012}.  To evaluate the accuracy of the network, the top-1 accuracy was computed on the GTSRB test set.

		\begin{table}[ht]
			\begin{center}
\vspace{1 cm}
				\caption{Top-1 accuracy results and number of parameters of MicronNet on German traffic sign recognition benchmark (GTSRB)~\cite{GTSRB}.  The results of several state-of-the-art traffic sign recognition are provided, along with the average human performance, for comparison purposes. }
				\label{Tab:res}
				\normalsize
				\begin{tabular}{|c|c||c|c|c|c|c|}
					\hline
					Model  & Number of  & Top-1 accuracy  \\
					Name & parameters & (GTSRB)   \\\hline \hline
					Human~\cite{humaAccuracy} & - & 98.8\% \\
					STDNN~\cite{ArcosGarcia2018} & 14M &99.7\% \\
					HLSGD~\cite{Jin2014} & 23.2M &99.6\% \\
					MCDNN~\cite{Ciresan2012} & 38.5M &99.5\% \\
					CDNN$^\dag$~\cite{Ciresan2012} & 1.54M &98.5\% \\
					MicronNet (fp16) & 0.51M & 98.9\% \\
					MicronNet (fixed16) & 0.51M & 98.0\% \\\hline
				\end{tabular}\\
$^\dag$average reported top-1 accuracy
			\end{center}
		\end{table}

		\subsection{Information density and NetScore}
		The model efficiency of the proposed MicronNet network and the state-of-the-art traffic sign recognition networks also being compared were assessed by means of its information density~\cite{Canziani} and NetScore~\cite{NetScore} as well to obtain a better understanding of the amount of relative performance a given deep neural network captured with respect to a fundamental building block.  More specifically, the information density ($D$) of a deep neural network $\mathcal{N}$ is defined as the performance of the deep neural network (denoted by $a(\mathcal{N})$) divided by the number of parameters needed for representing it (denoted by $p(\mathcal{N})$),
		
		\begin{equation}
		D(\mathcal{N}) = \frac{a(\mathcal{N})}{p(\mathcal{N})}
		\end{equation}

By taking into account both model size and network performance by means of a single metric, information density (expressed as percent of top-1 accuracy per parameter in this study) provides a good representation of the network's ability to utilize its full modeling capacity, with higher information density indicating better network efficiency.

One aspect that information capacity does not account for is the computational cost for performing inference with a given deep neural network, which is important for real-time embedded applications.  Therefore, the NetScore~\cite{NetScore} metric was also leveraged in this study for assessing the performance of a deep neural network $\mathcal{N}$ for practical usage.  The NetScore metric (denoted here as $\Omega$) can be defined as:
		\vspace{-0.2cm }
		\begin{equation}
		\Omega(\mathcal{N}) = 20\log\left(\frac{{a(\mathcal{N})}^{\alpha}}{{p(\mathcal{N})}^{\beta}{m(\mathcal{N})}^{\gamma}}\right)
		\end{equation}
\vspace{-0.2cm }

\noindent where $a(\mathcal{N})$ is the accuracy of the network, $p(\mathcal{N})$ is the number of parameters in the network, $m(\mathcal{N})$ is the number of multiply–accumulate (MAC) operations performed during network inference, and $\alpha$, $\beta$, $\gamma$ are coefficients that control the influence of accuracy, architectural complexity, and computational complexity of the network on $\Omega$.  We set $\alpha=2$, $\beta=0.5$, and $\gamma=0.5$ as proposed in~\cite{NetScore}.

		\begin{table}[ht]
			\begin{center}
\vspace{1 cm}
				\caption{Information density of MicronNet on German traffic sign recognition benchmark (GTSRB)~\cite{GTSRB}.  The results of several state-of-the-art traffic sign recognition networks are provided for comparison purposes.  Higher is better.}
				\label{fig:infodensity}
				\normalsize
				\begin{tabular}{|c||c|c|c|c|c|c|}
					\hline
					Model  & Information capacity  \\
					Name & (\% per MParams)   \\\hline \hline
					STDNN~\cite{ArcosGarcia2018} & 7.1 \\
					HLSGD~\cite{Jin2014} & 4.3 \\
					MCDNN~\cite{Ciresan2012} & 2.6 \\
					CDNN~\cite{Ciresan2012} & 64 \\
					MicronNet (fp16) & 194 \\
					MicronNet (fixed16) & 192 \\\hline
				\end{tabular}\\
			\end{center}
		\end{table}

		\subsection{Discussion}
\noindent \textbf{Top-1 accuracy.}		Table~\ref{Tab:res} shows the number of parameters and the top-1 accuracy of the proposed MicronNet network (both in half-precision floating-point data representation and 16-bit fixed-point data representation) on the GTSRB test dataset, along with the number of parameters and top-1 accuracies for state-of-the-art traffic sign recognition networks.  A number of interesting observations can be made.  First, the resulting MicronNet possesses just $\sim$510,000 parameters, which is $\sim$27.5x fewer than the state-of-the-art STDNN network~\cite{ArcosGarcia2018}.  Even when compared to the smallest state-of-the-art traffic sign recognition network compared in this paper (i.e., the CDNN network~\cite{Ciresan2012}, which MicronNet outperforms), the proposed MicronNet network still has $\sim$3x fewer parameters.  The significantly smaller number of parameters in the proposed MicronNet network compared to all of the evaluated state-of-the-art traffic sign recognition networks illustrates its efficacy for greatly reducing the computational and memory requirements, making the use of MicronNet very well suited for real-time embedded traffic sign recognition purposes.  Second, it can be observed that the resulting MicronNet was still able to achieve a top-1 accuracy of 98.9\% on the GTSRB test dataset, which is just $\sim$0.8\% lower than that achieved using the state-of-the-art STDNN network, and $\sim$0.4\% higher than that achieved by the smallest tested network outside of the proposed MicronNet (i.e., CDNN~\cite{Ciresan2012}).  Third, it can be observed that the top-1 accuracy of the proposed MicronNet network was equivalent to the average human performance reported in~\cite{humaAccuracy}.  The top-1 accuracy results exhibited by MicronNet illustrates the efficacy of this proposed network for providing strong embedded traffic sign recognition capabilities despite its significantly smaller size compared to other state-of-the-art networks.  In addition, it can be observed that the variant of the proposed MicronNet with 16-bit fixed-point data representation, while achieving lower top-1 accuracy than the proposed MicronNet with half-precision data representation, still managed to achieve a top-1 accuracy of 98.0\% on the GTSRB test dataset.

\begin{figure}[t]
			\begin{center}
				\includegraphics[width = 8 cm]{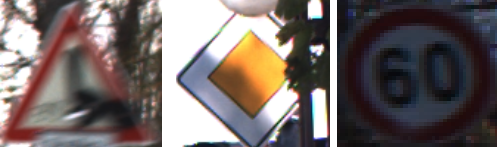}
			\end{center}
			\caption{Examples of traffic images from the GTSRB test dataset that has been misclassified by the proposed MicronNet.  It can be seen that in the example misclassified traffic images, the sign is either heavily motion blurred (left), partially occluded (middle), or exhibit poor illumination (right).}
			\label{fig:misclass}
		\end{figure}

To study where the proposed MicronNet encounters difficulties, we examine some of the traffic images from the GTSRB test dataset that has been misclassified by the proposed MicronNet (see Fig.~\ref{fig:misclass}).  It can be observed that in the example misclassified traffic images, the sign is either heavily motion blurred (left), partially occluded (middle), or exhibit poor illumination (right).  The identification of such misclassifications can provide good insight into the weaknesses of a network, as one potential mechanism for improving the robustness to such scenarios may be to extend the data augmentation policy to include more synthetic examples at different forms of occlusions as well as different illumination levels.

		~\\~\\
\noindent \textbf{Information density and NetScore.}		Table~\ref{fig:infodensity} shows the information density of the proposed MicronNet network on the GTSRB test dataset, along with the information density for state-of-the-art traffic sign recognition networks.  It can be observed that the information density of the resulting MicronNet is significantly higher than all of the other tested traffic sign recognition networks, by as much as $\sim$75x higher in the case of MCDNN~\cite{Ciresan2012}.  The high information density of the proposed MicronNet network, for both half-precision floating-point and 16-bit fixed-point data representations, when compared to the other evaluated state-of-the-art traffic sign recognition networks further illustrate the network efficiency of the proposed network.  Finally, the NetScore of the proposed MicronNet network was computed to be 102.52, which is quite high and further reinforces the strong balance between accuracy, architectural complexity, and computational cost of the proposed network.

		~\\~\\
\noindent \textbf{Resource usage.}				Table~\ref{Tab:macs-runtime} shows the resource usage of the proposed MicronNet network, which is very important for evaluating its efficacy for real-time embedded applications given that both memory and computational resources are very limited in such cases.  A number of interesting observations can be made.  First, it can be observed that the proposed MicronNet network is just $\sim$1.05MB in size, which can be contributed to the fact that not only is the number of parameters being very low compared to existing state-of-the-art networks but also a result of the fact that the parameters are represented with half-precision float-point values.  Second, it can be observed that the proposed MicronNet network requires just $\sim$10.5 million multiply-accumulate (MAC) operations to perform inference, which indicates that the proposed MicronNet network has low computational requirements for performance network inference.  To better evaluate the computational requirements of the proposed MicronNet network in a real-world embedded scenario, the network was evaluated on a 1.2GHz Cortex-A53 high efficiency processor in a Broadcom BCM2837B0 SoC.  It was found that the time-to-compute was just 32.19 ms on the tested high efficiency processor in half-precision floating-point (fp16) mode with power consumption of $\sim$3W, making it very well-suited for real-time embedded traffic sign recognition.  These experimental results clearly demonstrate that very small yet accurate deep neural network architectures can be designed for real-time traffic sign recognition that are well-suited for embedded scenarios.
			~\\~\\	
\noindent \textbf{Robustness against image degradation.} To study the robustness of the proposed MicronNet network against different levels of image degradation, all 12,630 images in the GTSRB test dataset were contaminated by Gaussian noise at three different degradation levels (i.e., $\sigma$= 2.5\%, 5\%, and 7.5\% of the dynamic range).  Table~\ref{Tab:degradation} shows the top-1 accuracy of the proposed MicronNet network across the different degradation levels.  It can be observed that the proposed MicronNet network is reasonably robustness to image degradation, still achieving a top-1 accuracy of 92.3\% at the highest tested degradation level ($\sigma$=7.5\%).

		\begin{table*}[ht]
			\begin{center}
\vspace{1 cm}
				\caption{Resource usage of MicronNet.  The time-to-compute was computed on a 1.2GHz Cortex-A53 high efficiency processor.}
				\label{Tab:macs-runtime}
				\normalsize
				\begin{tabular}{|c|c|c|c|c|}
					\hline
					Model  &Total number & Total number & Time To & Power\\
					Size &of Parameters &of MACs & Compute (fp16) & Consumption (W) \\\hline \hline
					1.05MB &0.51M & 10.5M & 32.19 ms & 3W\\\hline
				\end{tabular}
			\end{center}
		\end{table*}

		\begin{table}[ht]
			\begin{center}
\vspace{1 cm}
				\caption{Robustness of MicronNet against different levels of image degradation.}
				\label{Tab:degradation}
				\normalsize
				\begin{tabular}{|c|c|c|c|c|}
					\hline
					Degradation level  & $\sigma$=0\% & $\sigma$=2.5\% & $\sigma$=5\% & $\sigma$=7.5\%\\\hline \hline
					Top-1 accuracy  & 98.9\% & 98.5\% & 96.5\% & 92.3\% \\\hline
				\end{tabular}
			\end{center}
		\end{table}

		\section{Conclusions}
		In this paper, a highly compact deep convolutional neural network called MicronNet is introduced for real-time embedded traffic sign recognition.  By designing a highly optimized network architecture where each layer's microarchitecture is optimized to have as few parameters as possible, along with macroarchitecture augmentation and parameter precision optimization, the resulting MicronNet network achieves a good balance between accuracy and model size as well as inference speed.  The resulting MicronNet possess a model size of just $\sim$1MB and $\sim$510,000 parameters ($\sim$27x fewer parameters than state-of-the-art), requires just $\sim$10 million multiply-accumulate operations to perform inference (with a time-to-compute of 32.19 ms on a Cortex-A53 high efficiency processor), while still achieving a top-1 accuracy of 98.9\% on the German traffic sign recognition benchmark, thus achieving human-level performance.  These experimental results show that very small yet accurate deep neural network architectures can be designed for real-time traffic sign recognition that are well-suited for embedded scenarios.

Future work involves exploring extensions upon MicronNet across a larger range of traffic datasets to improve generalizability in different scenarios.  Furthermore, it is also worth exploring and investigating this integrated microarchitecture-level and macroarchitecture-level design principles and optimization strategies on deep neural network architectures for different tasks outside of traffic sign recognition, and the fundamental tradeoffs between microarchitecture-level and macroarchitecture-level design principles and optimization strategies on such deep neural network architectures and mechanisms to optimize for such tradeoffs to improve generalizability of such an approach.  Furthermore, model stability studies that also involve assessing the performance of this approach in the case of less training data given smaller model sizes would be quite interesting to explore as future work.  Finally, model performance studies with a wider variety of embedded processors at different floating-point and fixed-point precision levels would be interesting to explore as future work.

\vspace{-0.35 cm}
\section*{Acknowledgment}
\vspace{-0.15 cm}
The authors thank the Natural Sciences and Engineering Research Council of Canada, Canada Research Chairs Program, and DarwinAI, as well as Nvidia for hardware support.

{\small
\bibliographystyle{plain}
\bibliography{ccn_style}
}

\end{document}